\documentclass[letterpaper]{article}
\usepackage{aaai}
\usepackage{times}
\usepackage{helvet}
\usepackage{courier}
\usepackage{graphicx}
\begin{document}
%
\title{On Chatbots Exhibiting Goal-Directed Autonomy in Dynamic Environments} 
\author{Biplav Srivastava\\
IBM Research\\
}
\maketitle
\begin{abstract}
\begin{quote}
Conversation interfaces (CIs), or chatbots, are a popular form of intelligent agents that engage humans in task-oriented or informal conversation. In this position paper and demonstration, we argue that chatbots working in dynamic environments, like with  sensor data, can not only serve as a promising platform to research issues at the intersection of learning, reasoning, representation and execution for goal-directed autonomy; but also handle non-trivial business applications. We explore the underlying issues in the context of {\em Water Advisor}, a preliminary multi-modal conversation system that can access and explain water quality data. 
\end{quote}
\end{abstract}
      

\section{Introduction}
\label{sec:intro}

Chatbots \cite{dialog-intro}, which can engage people in natural dialog conversation, have gained popularity recently drawn by numerous platforms to create them quickly for any domain \cite{chatbot-survey-accenture}. Most common types of such agents deal with a single user at a time and conduct informal conversation, answer the user's questions or provide recommendations in a given domain.
They need to handle uncertainties related to human behavior and natural language, while conducting dialogs to achieve system goals. Chatbots have been deployed in customer care in many industries where they are expected to save over \$8 billion per annum by 2022 \cite{chatbot-cc-juniper}. 

However, the data sources used by common chatbots are static databases like product catalogs or user manuals. Therefore, for their problem of dialog management, i.e., creating dialog responses to user's utterances,   effective approaches include learning policies over predictable nature of data\cite{young2013pomdp} or  reasoning on its abstract representations \cite{minim-dialog}. 

The application scenarios become more  compelling when the chatbot works in a dynamic environment, e.g., with sensor data, and interacts with groups of people, who come and go, rather than only an individual at a time. In such situations, the agent has to execute actions to monitor the environment, model different users engaged in conversation over time and track their intents, learn patterns and represent them, reason about best course of action given goals and system state, and execute conversation or other multi-modal actions. 

We now explore the underlying issues of goal-directed autonomy in dynamic environment in the context of {\em Water Advisor} (WA) \cite{wa-aaai18}, a prototypical multi-modal conversation system that can access and explain water quality data
to a variety of stake-holders. We identify opportunities for 
learning, reasoning, representation and execution in WA and motivate more such applications.

\section{Decision-Support for Water Usage With a Multi-Modal Conversation Interface}
\label{sec:water-adv}

The global situation of water quality around the world is alarming in both developing and developed countries\cite{un-water} because water demand continues to rise while existing sources for fresh water are getting polluted. A key strategy for tackling  water pollution is engaging people. 
A person makes many daily decisions touching on water usage activities like for
 profession (e.g., fishing, irrigation, shipping), recreation (e.g, boating), wild life conservation (e.g., dolphins) or just regular living (e.g., drinking, bathing, washing). Accessible tools for public are particularly useful to handle public health challenges such as the Flint water crisis \cite{flint}. 
 
 A decision in this space needs to consider the activity (purpose) of the water use; relevant water quality parameters and their applicable regulatory standards for safety; available measurement technology, process, skills and costs; and actual data. 
 There are further complication factors: there may be overlapping regulations due to geography and administrative scope; one may have to account for alternative ways to measure a particular water quality parameter that evolves over time; and water data can have issues like missing values or at different levels of granularity. 
 The very few tools available today target water experts such as WaterLive mobile app for Australia \footnote{http://www.water.nsw.gov.au/realtime-data
},  Bath app for UK\footnote{2https://environment.data.gov.uk/bwq/profiles/}, and GangaWatch for India \cite{gw-demo} and assume technical understanding of sciences. 



\begin{figure*}
 \centering
   \includegraphics[width=0.7\textwidth]{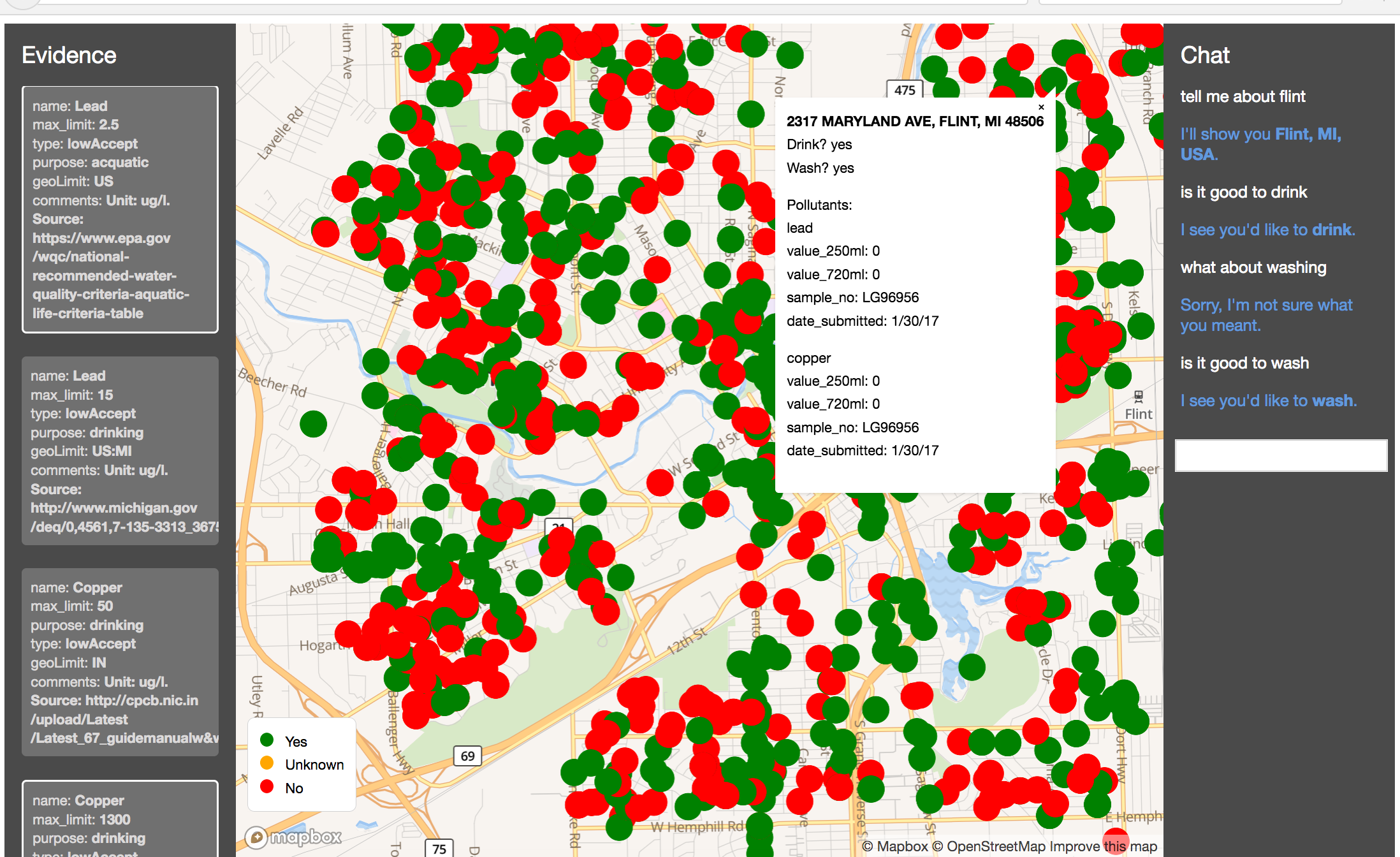}
  \caption{A screenshot of {\em Water Advisor}. See video of it in action at https://youtu.be/z4x44sxC3zA.}
  \label{fig:screen-shot-wa}
\end{figure*}


\subsection{Water Advisor}

Water Advisor (WA) is intended to be a  data-driven assistant that can guide people without requiring any special water expertise.  One can trigger it via a conversation to get an overview of water condition at a location, explore it by filtering and zooming on a map, and seek details on demand (Figure~\ref{fig:screen-shot-wa}) by exploring relevant regulations, data or other locations.
The current prototype uses water quality data available from Flint, MI\footnote{http://www.michigan.gov/flintwater/0,6092,7-345-76292\_76294\_76297---,00.html}
but future extensions will use open water data from US Geological Survey\footnote{https://waterdata.usgs.gov/nwis/current/?type=quality} (USGS) that is refreshed for thousands of places in US per day. However, the number of water quality parameters, for which data is available, varies widely between locations and over time, making generation of useful advice challenging. 
For regulations, WA relies on information provided by multiple agencies at national (US, India) and state levels (Michigan, New York), which has been consolidated for reuse\footnote{https://github.com/biplav-s/water-info}.

\subsection{Technical Issues}


In a water advising application, one or more users may need to interact with the chatbot if handling a complex decision like water contamination. 
The tool has to detect the user's information goals and meet them at lowest cognitive cost. The system uses a natural language classifier (NLC) to understand user utterance, and its error rate varies with input. The system has to decide whether to ask clarifying questions if it has low confidence and there are many ways to respond. The user may have preferences about how they specify an input (like location) and the kind of response 
they want (visual v/s textual). We discuss a range of issues below for exposition {\em but note that the current WA prototype handles only a subset of following integration issues}. 


\subsubsection{Learning} plays an important role in understanding user's utterance, finding reliable water data samples in the database based on region and duration of interest, discovering issues in water quality and improving overall performance over time. In the prototype, for utterances, we use trained user models from commercial systems and for water quality, a simple regression method.

\subsubsection{Representation} is needed to map water's usage purpose to quality parameters  and model safe limits of pollution parameters with different mathematic properties (e.g., polarity). It also helps map water purpose to regulations and further, aggregate and reconcile the latter when a region falls under overlapping jurisdiction of regulations. We represent this as geographically-scoped attribute-value pair in JSON format and make it publicly available for others to use and extend\footnote{https://github.com/biplav-s/water-info}. 

\subsubsection{Reasoning} is crucial to keep conversation focused based on system usability goals and user needs.  One can model cognitive costs to user based on alternative system response choices and seek to optimize short-term and long-term behavior. Reasoning can further help to short-list regulations based on water activity and region of interest, generate advice and track explanations. We currently use rules on geographical scope and missing values to determine system response.

\subsubsection{Execution} is autonomous as the agent can choose to act
by (a) asking clarifying questions about water usage goals or locations, (b) asking user's preference
about advice, (c) seeking most reliable water data for region and time interval of interest from available external data sources, and corresponding subset of compatible regulations (d) invoking reasoning to generate an advice for water usage using filtered water data and regulations, (e) visualizing and explaining its output using water regulations, and (f) using one or more suitable modalities available at any turn of user interaction, i.e., chat, maps and document views. 
The current prototype uses a simplistic  strategy for execution based on error rates, system confidence and usability rules.

\subsubsection{Human Usability Factors} have to be modeled and supported during WA's operation. In the current prototype, the user-interface controller module automatically keeps the different modalities synchronized so that the user is looking at consistent information across them. The system has to be aware of missing data or assumptions it is making, and needs to take them into account while communicating output advice in generated natural language.  One avenue for future exploration is to measure and track complexity of interaction \cite{dial-comp} and use sensed signals to pro-actively improve user experience. Another is to combine close-ended and open-ended questioning strategies for efficient interaction \cite{open-close-iui18}.

\subsubsection{Ethical Issues} can emerge whenever a piece of technology is used among people at large. In the context of conversations, a recent paper surveys ethical issues \cite{dialog-ethics} like biases, adversarial
examples, privacy violations, safety
challenges and reproducibility concerns. 
A water-use chatbot can conceivably create bias among users of different activity subgroups (e.g., preferring recreation over drinking), compromise on privacy of users who submit queries about an activity or a region, and create public safety concerns (e.g., when users find scarcity of good quality water). We have not considered them in the prototype, however.




\section{Discussion and Conclusion}
\label{sec:conc}

In this paper, we used decision-support in water as a use-case to demonstrate  that chatbots can serve as a promising platform to integrate AI sub-disciplines for goal-directed autonomy.   Apart from learning, reasoning, representation and execution, chatbots also need to work with  human usability factors and ethical issues.
An interesting aspect of these applications is that the chatbot may be helping a group of people take collective decision making, like conducting an interview, and data changes over time.
Beyond water and customer support, complex
applications are emerging in sciences (astronomy\cite{tyson-aaai18}), business (career counseling\footnote{https://www.ibm.com/talent-management/career-coach}, hospitality\footnote{https://www.bebot.io/hotels}) and societal domains (health\footnote{https://www.healthtap.com/}).





\bibliographystyle{named}
\bibliography{references}

\end{document}